\documentclass[conference]{IEEEtran}
\usepackage{cite}

\ifCLASSINFOpdf
\else
\fi
\usepackage{times}
\usepackage{epsfig}
\usepackage{graphicx}
\usepackage{amsmath}
\usepackage{amssymb}
\usepackage{multirow}
\usepackage[pagebackref=true,breaklinks=true,letterpaper=true,colorlinks,bookmarks=false]{hyperref}

\usepackage{fixltx2e}

\usepackage{stfloats}
\usepackage{etoolbox}
\begin{document}
%
\title{Proactive Human-Robot Interaction using Visuo-Lingual Transformers}

\author{\IEEEauthorblockN{Pranay Mathur} 
\IEEEauthorblockA{ Georgia Institute of Technology}
}

\maketitle

\begin{abstract}
Humans possess the innate ability to extract latent visuo-lingual cues to infer context through human interaction. During collaboration, this enables proactive prediction of the underlying intention of a series of tasks. In contrast, robotic agents collaborating with humans naively follow elementary instructions to complete tasks or use specific hand-crafted triggers to initiate proactive collaboration when working towards the completion of a goal. Endowing such robots with the ability to reason about the end goal and proactively suggest intermediate tasks will engender a much more intuitive method for human-robot collaboration. To this end, we propose a learning-based method that uses visual cues from the scene, lingual commands from a user and knowledge of prior object-object interaction to identify and proactively predict the underlying goal the user intends to achieve. Specifically, we propose ViLing-MMT, a vision-language multimodal transformer-based architecture that captures inter and intra-modal dependencies to provide accurate scene descriptions and proactively suggest tasks where applicable. We evaluate our proposed model in simulation and real-world scenarios.
\end{abstract}

%
\IEEEpeerreviewmaketitle

\section{Introduction}
Robots have found increasing application in working with humans in diverse spaces including manufacturing, healthcare and recreation. This can be attributed to improved levels of safety \cite{haddadin2012making} and an accurate understanding of user intentions \cite{wang2013probabilistic}. Advances in multi-modal learning combining vision and linguistic knowledge have enabled the use of semantic contexts to improve the accuracy of inferring user intention. Utilizing multi-modal learning for efficient human-robot collaboration has proven to be a promising research direction and multiple methods incorporating this have been proposed \cite{askingforhelp,vision_language_navigation}. 
Typical approaches in human-robot collaboration attempt to hand-craft interaction logic\cite{hand-craftHRI}. Newer methods use data-driven approaches to learn responses or reactions to human actions \cite{datadrivenHRI,crowdsourcing_hri,learning_hri_gameplay,learning_social}. A drawback of the methods above is the requirement for manual extension of the interaction logic to new tasks and heavy dependence on hand-crafted logic for triggering proactive behaviour making it unscalable.

Teaching a robot pro-active behaviour in the absence of explicitly modelled user intention is a non-trivial task \cite{nov_prosoc,prosocrobo}. Our multi-modal model architecture infers user intention and automatically suggests an intermediate goal. Our work specifically addresses the scalability of learning new tasks and ensuring that no explicitly well-defined triggers are needed to initiate proactive behaviour. Specifically, the contributions of this work are two-fold:\\
1) An end-to-end multimodal transformer architecture that uses visual cues from the scene and intermediate task instructions to initiate pro-active behaviour \\
2) Incorporating graphical representation of learnt prior object-object relations in an unsupervised manner 

We demonstrate the success of our model in simulation and in real-world scenarios and analyze our results in Section \ref{results}. 

\section{Methodology}
\begin{figure*}[!ht]
\begin{center}
   \includegraphics[width=\textwidth]{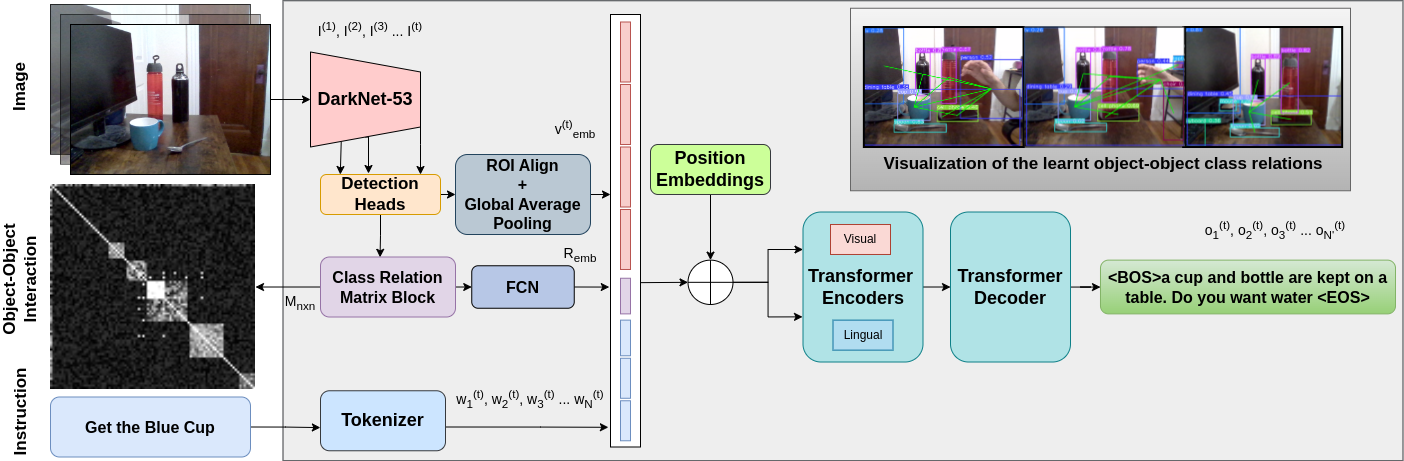}
\end{center}
   \caption{Proposed Architecture and Example Task: The input images, the visualized object-object interaction graph, and the instruction passed by the user. A visualization of class relations learnt can be seen in the upper-right corner. The output sequence describes the scene and identifies user intention accurately}
\label{fig:model_arch}
\vspace{-12pt}
\end{figure*}
We present our architecture visuo-lingual multimodal transformers (ViLing-MMT) to generate scene descriptions and proactively provide task suggestions in Figure \ref{fig:model_arch}.
 
\subsection{Vision Encoder}
To incorporate visual context awareness from an image sequence $I^{(1)}, I^{(2)}, I^{(3)}\dots ,I^{(t)}$, we create a visual embedding $\textbf{v}^{(t)}_{emb}$ using an encoder based upon the Darknet-53 neural network architecture \cite{redmon2018yolov3}. We generate image region features by extracting bounding-boxes and their visual features. As they are of different sizes, we apply RoIAlign pooling to normalize the sizes of feature maps as well as global average pooling (GAP) to reduce the feature representation dimension. Through this process, we are able to represent each class of object detected in the image as a 512-dimensional vector. The 2D position embedding vector is appended to this vector along with the embedding vector from the class relation matrix block and tokenized lingual commands.

\subsection{Object Relation Encoder}
The output from Darknet-53 is branched and sent to the detection heads where we use class occurrences of objects to form a graph encoding historical object-object relations. The weighted graph $G(V, E)$ is represented as a symmetric adjacency matrix $M_{n\times n}$ for $n$ classes. Each class $c_n$ is represented as a vertex $v_{cn} \in V$, $N(V)=n$ and a relation is denoted by an edge $e_{c_1c_2}$ between two vertices $V_{c_1}, V_{c_2} \in V $. The weight $w_{c_1c_2}$ of the edge $e_{c_1c_2}$ is a measure of the extent to which the object classes $c_1-c_2$ are related. We illustrate our method to quantify this object-object relation between classes $c_1-c_2$ for each $c_1,c_2 \in C_n$ where $C_n$ is the list of classes detected -

\begin{equation}
    w(c_1,c_2)=\frac{ N(c_1 \cap c_2) }{N(c_1)\cdot N(c_2)}
\end{equation}

where $N(c_1 \cap c_2)$ is the number of times the classes occur in the same frame, and $N(c_1), N(c_2)$ denote a count of their individual appearances. The matrix $M_{n\times n}$ is then flattened and passed through a class relation matrix encoder resulting in an embedding $\textbf{R}_{emb}$.

\subsection{Transformer Encoder}
The instructions to the robot are represented by a sequence of words. The input sequence is split into multiple tokens $w^{(t)}_1, w^{(t)}_2, w^{(t)}_3 \dots w^{(t)}_N$, along with the vision embeddings $\textbf{v}^{(t)}_{emb}$ discussed above, which are used to create dense vision-language embeddings $\textbf{vRL}^{(t)}_{emb}$ using a transformer encoder\cite{vaswani2017attention}. While images used spatial encoding as position embeddings, standard positional encoding proposed in \cite{vaswani2017attention} is used for textual inputs. Our encoder architecture shares architectural similarities with ViLBERT\cite{lu2019vilbert} inspired by BERT (Bidirectional Encoder Representations from Transformers) \cite{devlin2018bert}. We incorporate the underlying idea of using two multi-modal streams of data consisting visual embeddings and language tokens that interact through their proposed co-attentional transformer layers. This permits variable individual modality-specific depths and promotes cross-modal connections at various depths. The model also computes the query, key, and value matrices as in a transformer, the novelty is that the keys and values are communicated to the attention block of the other modality. This results in cross-modality attention-pooled features. 
 
\subsection{Transformer Decoder}
We use a transformer decoder\cite{vaswani2017attention} to process the fixed-length context vector $\textbf{vRL}^{(t)}_{emb}$. The transformer-based decoder defines the conditional probability distribution of the target sequence $\textbf{o}_{1:N'}$ given the contextualized encoding sequence
\begin{flalign}
\begin{aligned}
       &\textbf{p}_{\theta_{enc},\theta_{dec}}(\textbf{o}_{1:N'}|\textbf{w}_{1:N},\textbf{I}^{(t)},M_{n\times n})\\
&= \prod_{i=1}^{N'}\textbf{p}_{\theta_{dec}}(\textbf{o}_{i}|\textbf{o}_{0:i-1}, \textbf{vRL}^{(t)}_{emb})  \forall i \in {1,\cdots,N'}  
\end{aligned}
\end{flalign}

The decoder consists of a stack of decoder blocks followed by a dense layer that maps the contextualized encoded sequence  $\textbf{vRL}^{(t)}_{emb}$ and the target vector sequence $\Bar{\textbf{o}}_{1:N'}$. Given intermediate visual and linguistic representations, the module computes query, key, and value matrices as in a standard transformer block similar to ViLBERT\cite{lu2019vilbert}.
\begin{figure}[!ht]
\includegraphics[width=0.5\textwidth]{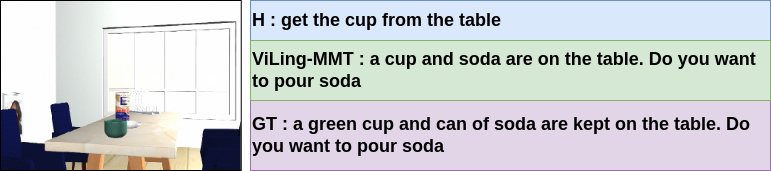}
\caption{Human instruction, model output description with triggered proactive response and ground truth reference caption in a simulated scene}
\label{fig: result}
\vspace{-10pt}
\end{figure}

\section{Results} \label{results}
\begin{table}[]
\caption{Results: Precision, Recall, F1 and BLEU score}
\vspace{-15pt}
\label{tab1}
\begin{center}
\begin{tabular}{|c|cccc|}
\hline
\multirow{2}{*}{\textbf{Model}} & \multicolumn{4}{c|}{\textbf{Simulated Scene}}                                                                               \\ \cline{2-5} 
                                & \multicolumn{1}{c|}{\textbf{Precision}} & \multicolumn{1}{c|}{\textbf{Recall}} & \multicolumn{1}{c|}{\textbf{F1}} & \textbf{BLEU} \\ \hline
\textbf{ViLing-MMT-G}           & \multicolumn{1}{c|}{0.625}          & \multicolumn{1}{c|}{0.667}         & \multicolumn{1}{c|}{0.645}       & 0.418         \\ \hline
\textbf{ViLing-MMT}             & \multicolumn{1}{c|}{0.867}          & \multicolumn{1}{c|}{0.813}         & \multicolumn{1}{c|}{0.838}       & 0.498         \\ \hline
\multirow{2}{*}{\textbf{Model}} & \multicolumn{4}{c|}{\textbf{Real-World Scene}}                                                                              \\ \cline{2-5} 
                                & \multicolumn{1}{c|}{\textbf{Precision}} & \multicolumn{1}{c|}{\textbf{Recall}} & \multicolumn{1}{c|}{\textbf{F1}} & \textbf{BLEU} \\ \hline
\textbf{ViLing-MMT-G}           & \multicolumn{1}{c|}{0.734}          & \multicolumn{1}{c|}{0.734}         & \multicolumn{1}{c|}{0.734}       & 0.526         \\ \hline
\textbf{ViLing-MMT}             & \multicolumn{1}{c|}{0.750}          & \multicolumn{1}{c|}{1}             & \multicolumn{1}{c|}{0.857}       & 0.566         \\ \hline
\end{tabular}
\end{center}
\vspace{-13pt}
\end{table}
We use the Flickr8K \cite{Flickr8k} and MSCOCO \cite{mscoco} datasets for pre-training the transformer encoder and visual encoder respectively. The Flickr8K dataset contains 8,000 images, each annotated with 5 reference captions. For end-to-end training, we customized the dataset, cherry-picking scenes where user intention is known and appending it to the description. 

Similar studies evaluate their approach using subjective evaluations by users', success rate of interaction initiation \cite{Liao2016,rashed2016observing} or the recognition accuracy of user intention \cite{shi2015measuring}. We evaluate our approach using precision and recall of the number of times proactive behaviour was triggered when expected and use BLEU score \cite{bleuscore} to evaluate the quality of scene descriptions and proactively suggested tasks. ViLing-MMT is evaluated both with and without the object-object interaction graph $G$ in simulation and the real world. Deteriorated performance was observed across all metrics when the graph $G$ was removed from the architecture (ViLing-MMT-G). An example is illustrated in Figure \ref{fig: result} and results have been summarized in Table \ref{tab1}. Our model successfully initiated proactive behaviour without hand-crafted triggers in both scenarios.

{\small
\bibliographystyle{ieee_fullname}
\bibliography{references}
}

\end{document}